\documentclass[letterpaper, 10 pt, conference]{ieeeconf}  
\usepackage{graphicx}
\usepackage{booktabs}

\IEEEoverridecommandlockouts                              

\usepackage{amsmath} 
\usepackage{amssymb}  
\usepackage{multirow}
\usepackage{makecell}
\usepackage{siunitx}

\title{\LARGE \bf
C2Dex: Contact-Consistent Reconstruction and Retargeting\\
for Dexterous Manipulation from Monocular Video
}

\author{%
Jie Ren$^{1,*}$,
Zhehao Jiang$^{1,*}$,
Yinhong Yang$^{1,*}$,
Haorui Jia$^{1}$,
Han Jiang$^{1}$,
Ben Li$^{2}$,
Yao Yao$^{1}$,\\
Cheng Lin$^{3}$,
Qiu Shen$^{1}$,
Zhenshan Bing$^{1}$,
Xiao-Xiao Long$^{1,\dagger}$,
Xun Cao$^{1}$%
\thanks{$^{*}$Equal contribution.}%
\thanks{$^{1}$Nanjing University, Nanjing, China.}%
\thanks{$^{2}$China Mobile Research Institute, Beijing, China.}%
\thanks{$^{3}$Macau University of Science and Technology, Macau, China.}%
\thanks{$^{\dagger}$Corresponding author: Xiao-Xiao Long
(\texttt{xxlong@nju.edu.cn}).}%
\thanks{This work has been submitted to the IEEE for possible publication.
Copyright may be transferred without notice, after which this version may
no longer be accessible.}%
}

\begin{document}

\maketitle
\thispagestyle{empty}
\pagestyle{empty}


\begin{abstract}
High-quality demonstrations for dexterous robot manipulation are costly and difficult to collect, whereas monocular human videos provide a scalable source of diverse manipulation behaviors. However, transferring such demonstrations to dexterous robots remains challenging: monocular hand--object interaction (HOI) reconstruction often produces temporally unstable contacts and physically implausible interactions, while conventional retargeting methods struggle to preserve task-relevant contacts and local interaction geometry across different hand embodiments. We present C2Dex, a video-to-dexterous-manipulation framework built around a shared interaction representation: stable object-side contacts recovered by aggregating noisy frame-wise observations in the canonical object space. These stable contacts serve a dual role: as trajectory-level constraints that guide reconstruction toward temporally coherent and physically plausible human HOI trajectories, and as explicit transfer targets for the dexterous hand, where Laplacian interaction optimization preserves the local hand--object geometry across embodiments and residual reinforcement learning refines the trajectory in simulation. Experiments on DexYCB and TACO show that C2Dex achieves end-to-end trajectory success rates of $57.78\%$ and $26.67\%$, respectively, substantially outperforming the strongest baselines ($17.78\%$ and $10.00\%$) under identical evaluation criteria. Real-robot replay experiments further demonstrate physical feasibility across diverse contact-rich manipulation tasks. Project page: \texttt{https://k-jie.github.io/C2Dex/}.
\end{abstract}

\textperiodcentered{}
Imitation Learning
\textperiodcentered{}
Deep Learning in Grasping and Manipulation

\section{Introduction}
Dexterous manipulation requires demonstrations that capture the coordinated
evolution of hand configuration, object motion, and contact over time.
Collecting such demonstrations directly on robotic hands, however, typically
relies on teleoperation or motion-capture systems that demand specialized
hardware and substantial task-specific effort, making it costly to scale
data collection across objects, tasks, and hand
embodiments~\cite{wang2024dexcap,arunachalam2022holo,zhao2023learning}.
In contrast, monocular human videos are abundant, easy to capture, and rich
in contact-rich manipulation behaviors, providing a scalable source of
demonstrations and motivating recent efforts to reconstruct hand--object
interactions (HOIs) from videos and transfer them to dexterous
hands~\cite{qin2022dexmvimitationlearningdexterous,mandikal2022dexviplearningdexterousgrasping,chen2025vividexlearningvisionbaseddexterous,xu2026demobot,paliwal2026doasido,mu2026deximit}.

\begin{figure*}[t]
    \centering
    \includegraphics[width=\textwidth]{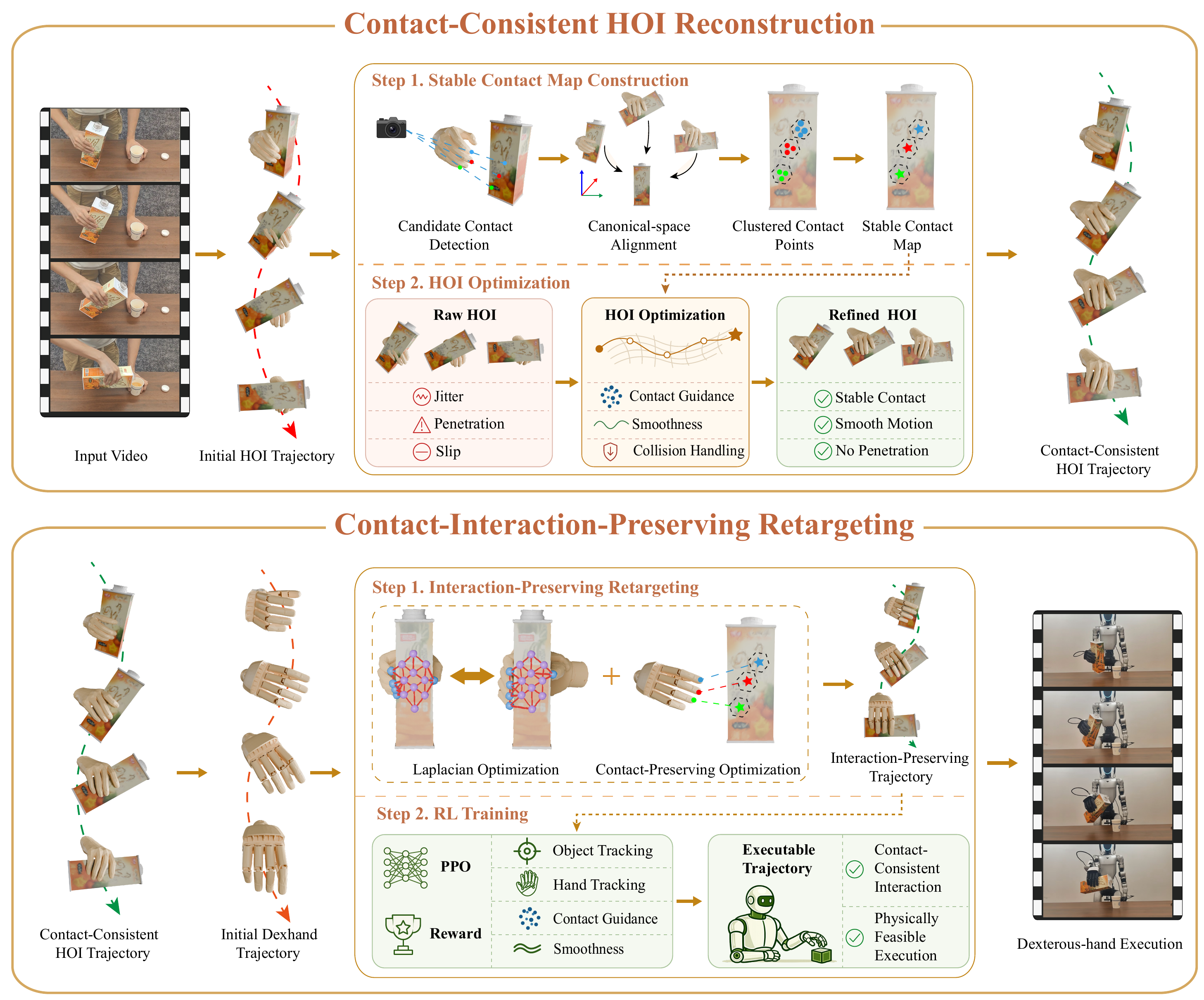}
    \caption{
        Overview of C2Dex. Given a monocular human video, C2Dex first performs
        contact-consistent HOI reconstruction by aggregating frame-wise contact
        observations into stable object-side contacts, and then transfers the
        reconstructed interaction to a dexterous hand through
        contact-interaction-preserving retargeting with stable-contact
        preservation, Laplacian interaction optimization, and residual
        reinforcement learning.
    }
    \label{fig:pipeline}
\end{figure*}

Generating executable dexterous-hand--object trajectories from monocular
videos remains challenging for two reasons. First, monocular HOI
reconstruction is highly sensitive to occlusion, pose drift, and depth
ambiguity, often yielding temporally inconsistent contacts, hand--object
penetration, and physically implausible trajectories. Second, even given an
accurate human trajectory, the morphological and kinematic gap between human
and dexterous hands makes it difficult to preserve task-relevant contacts
and local interaction geometry during retargeting. These errors accumulate
throughout the video-to-robot pipeline, and both failure modes share a
common root cause: the lack of a reliable, temporally stable description of
where the demonstrated interaction occurs on the object.

To address these challenges, we present \textbf{C2Dex}, whose key idea is to
recover \emph{stable object-side contacts} in the canonical object space and
share them across the entire video-to-robot pipeline. Aggregated over
locally stable temporal segments, this representation is robust to per-frame
noise and, being anchored to the object rather than any particular hand,
transfers naturally across embodiments. Two coupled modules consume it.
\emph{Contact-consistent HOI reconstruction} uses the stable contacts as
trajectory-level constraints to refine the human HOI trajectory, reducing
contact jitter, spurious contact switching, and penetration.
\emph{Contact-interaction-preserving retargeting} transfers the same
contacts to the dexterous hand as explicit targets and applies Laplacian
interaction optimization to preserve local hand--object geometry; collision
avoidance, joint limits, and temporal regularization ensure kinematic
feasibility, and a residual-RL stage~\cite{li2025maniptrans} further ensures
physical executability.

We evaluate C2Dex on end-to-end trajectory generation, HOI reconstruction,
retargeting, and real-robot replay. C2Dex achieves trajectory success rates
of $57.78\%$ on DexYCB and $26.67\%$ on TACO, versus $17.78\%$ and $10.00\%$
for the strongest baselines under identical criteria, while yielding more
physically stable reconstructions and more accurate contact transfer with
far less penetration. Real-robot replay on $24$ demonstrations across $8$
contact-rich tasks demonstrates the physical feasibility of the generated
trajectories.

Our main contributions are summarized as follows:
\begin{enumerate}
\item We propose a \textbf{stable object-side contact representation},
recovered by aggregating noisy frame-wise observations in the canonical
object space, serving as a shared interface between monocular HOI
reconstruction and cross-embodiment retargeting.

\item Building on this representation, we develop
\textbf{contact-consistent HOI reconstruction} for temporally coherent,
physically plausible HOI trajectories, and
\textbf{contact-interaction-preserving retargeting} with Laplacian
interaction optimization to preserve task-relevant contacts and local
hand--object geometry across embodiments.

\item We build the \textbf{video-to-dexterous-manipulation system} and
validate it on DexYCB, TACO, and a real robotic platform, substantially
outperforming existing video-to-dexterity pipelines in end-to-end
trajectory generation.
\end{enumerate}

\section{Related Work}
\subsection{Hand--Object Reconstruction from Monocular Video}
Building on advances in monocular hand
reconstruction~\cite{pavlakos2023reconstructinghands3dtransformers,ye2025predicting4dhandtrajectory},
template-free methods~\cite{fan2023holdcategoryagnostic3dreconstruction,
ye2023diffusionguidedreconstructioneverydayhandobject,
on2025bigsbimanualcategoryagnosticinteraction} jointly recover hand and
object geometry without a pre-scanned model, and recent
work~\cite{Liu_2025, wang2025magichoileveraging3dpriors,
shi2026agilehandobjectinteractionreconstruction} further exploits
large-scale geometric or generative priors for in-the-wild robustness.
However, these methods resolve contact independently per frame, so the
inferred contacts drift over time---tolerable for visualization but
unreliable for manipulation. C2Dex instead aggregates multi-frame contact
evidence in the canonical object space to obtain temporally consistent,
physically plausible interactions.

\subsection{Dexterous-Hand Retargeting}
Retargeting transfers human hand motion to a robotic hand with a different
kinematic structure, classically by matching fingertips or task-space
vectors~\cite{handa2020dexpilot, qin2023anyteleop}, and more recently through
learned geometric correspondences~\cite{yin2025geort,
lin2025dexflowunifiedapproachdexterous}. These objectives are defined purely on
hand-side similarity and assume accurate input trajectories, so residual
reconstruction errors propagate to the robot as penetration or lost contact.
C2Dex instead preserves interaction by transferring stable object-side contacts
and maintaining local hand--object geometry via Laplacian optimization.

\subsection{Learning Dexterous Manipulation from Human Videos}
Early methods extract motion or visual priors from human
demonstrations~\cite{qin2022dexmvimitationlearningdexterous,
mandikal2022dexviplearningdexterousgrasping}, while
ManipTrans~\cite{li2025maniptrans} and
DexMachina~\cite{mandi2025dexmachinafunctionalretargetingbimanual} transfer
captured HOI trajectories to dexterous hands via RL, assuming clean trajectories
are available. Recent systems~\cite{li2024okami,
kareer2024egomimicscalingimitationlearning,
hsieh2025dexmanlearningbimanualdexterous, paliwal2026doasido, mu2026deximit, zhou2025you,park2025learning}
drop this assumption and operate directly on monocular video, but inherit the
errors of the upstream perception stack. C2Dex targets these errors, emphasizing
temporally stable contacts and cross-embodiment interaction preservation.

\section{Method}
As illustrated in Fig.~\ref{fig:pipeline}, C2Dex converts a monocular human
video into an executable dexterous manipulation trajectory through two
tightly coupled modules. Contact-consistent HOI reconstruction
(Sec.~\ref{sec:hoi_recon}) recovers stable object-side contacts from noisy
frame-wise observations and uses them to refine the human HOI trajectory.
Contact-interaction-preserving retargeting (Sec.~\ref{sec:retargeting}) transfers
the reconstructed interaction to the target dexterous hand while preserving
task-relevant contacts and local hand--object geometry.

\subsection{Contact-Consistent HOI Reconstruction}
\label{sec:hoi_recon}

Given a monocular video $V=\{I_t\}_{t=1}^{T}$, we first estimate an initial
human HOI trajectory and extract frame-wise contact observations between the
hand and object, which are then stabilized and used to refine the hand motion
as follows.

\paragraph{Hand--Object State Initialization.}
Using Dyn-HaMR~\cite{yu2025dynhamrrecovering4dinteracting}, we estimate the
initial per-frame MANO~\cite{romero2022embodied} articulation
$\boldsymbol{\theta}^{0}_t$ and global hand transformation
$\boldsymbol{\tau}^{0}_t\in SE(3)$, together with a sequence-level shape
parameter $\boldsymbol{\beta}$. The resulting hand mesh is denoted by
$\mathcal{M}^{h,0}_t=(\mathcal{V}^{h,0}_t,\mathcal{F}^{h})$, where
$\mathbf{v}^{h,0}_{t,i}$ is its $i$-th vertex. We further use SAM
3D~\cite{sam3dteam2025sam3d3dfyimages} to reconstruct a canonical object mesh
$\mathcal{M}_o$ and estimate its per-frame 6D pose
$(\mathbf{R}_{o,t},\mathbf{p}_{o,t})$ using
ProxyPose~\cite{zhang2026proxypose}.

\paragraph{Cross-Frame Contact Stabilization.}
Contacts extracted independently from individual frames are sensitive to pose
noise, occlusion, and monocular depth ambiguity. Our key observation is that,
within a locally stable contact phase, a hand vertex should remain associated
with the same local region of the object surface. We therefore aggregate
frame-wise contact observations in the canonical object space.

For each frame, we render the initial hand mesh and the posed object mesh under
the estimated camera. Let $S_{h,t}$ and $S_{o,t}$ denote their image-space
silhouettes; their overlap $S_{c,t}=S_{h,t}\cap S_{o,t}$ defines a candidate
contact region.
We project the hand vertices onto the image plane and retain those whose
projections lie inside $S_{c,t}$. For each candidate vertex $i$, we cast a
ray from the camera center through its projected position and intersect it
with the reconstructed object surface. The resulting surface point
$\mathbf{x}_{t,i}$ is treated as the frame-wise object-side contact
observation associated with $\mathbf{v}^{h,0}_{t,i}$.

Silhouette overlap may also include vertices that occlude the object without
physically contacting it. We therefore filter the candidate pairs according
to surface-normal compatibility. Let $\mathbf{n}^{h}_{t,i}$ and
$\mathbf{n}^{o}_{t,i}$ denote the outward hand and object normals at
$\mathbf{v}^{h,0}_{t,i}$ and $\mathbf{x}_{t,i}$, respectively. We define the
normal-consistency score
$w^{n}_{t,i}
=
-\left(\mathbf{n}^{h}_{t,i}\right)^{\top}\mathbf{n}^{o}_{t,i}
/
\left(
\|\mathbf{n}^{h}_{t,i}\|_2
\|\mathbf{n}^{o}_{t,i}\|_2
\right)$,
which assigns a high score to approximately opposite surface normals. We
retain pairs satisfying $w^{n}_{t,i}>\gamma_n$ and denote the corresponding
hand-vertex indices by $\mathcal{C}_t$.

To compare the retained contact observations across frames, we transform each
$\mathbf{x}_{t,i}$ into the canonical object frame using the inverse object
pose and denote the result by $\bar{\mathbf{x}}_{t,i}$. This removes the rigid
object motion and enables direct comparison of contacts associated with the
same object-surface region.

We partition the sequence into locally stable temporal segments.
Two adjacent frames are assigned to the same segment when
$\|\boldsymbol{\theta}^{0}_{t+1}-\boldsymbol{\theta}^{0}_{t}\|_2
<\epsilon_{\theta}$
and the relative hand--object pose remains nearly unchanged,
where $\epsilon_{\theta}$ is the articulation-variation threshold.

Applying this criterion sequentially partitions the sequence into maximal
contiguous segments $\mathcal{S}_1,\ldots,\mathcal{S}_K$, and $k(t)$ denotes
the segment containing frame $t$.
For each hand vertex $i$, we collect its canonical contact observations
within segment $\mathcal{S}_k$. We cluster these observations using
DBSCAN~\cite{ester1996density} and
retain the dominant cluster, denoted by $\mathcal{X}_{k,i}$. If no valid cluster is found, $\mathcal{X}_{k,i}$ is set to $\varnothing$. The stable contact vertices in segment
$\mathcal{S}_k$ are then given by $
\widehat{\mathcal{C}}_k
=
\left\{
i
\;\middle|\;
\mathcal{X}_{k,i}\neq\varnothing
\right\}
$. Then,
for every $i \in \widehat{\mathcal{C}}_k$, we compute the stable canonical
contact point as the medoid of the dominant cluster:
\begin{equation}
\bar{\mathbf{x}}^{\mathrm{stab}}_{k,i}
= \operatorname*{arg\,min}_{\mathbf{x} \in \mathcal{X}_{k,i}}
  \sum_{\mathbf{y} \in \mathcal{X}_{k,i}} \|\mathbf{x} - \mathbf{y}\|_2,
\qquad i \in \widehat{\mathcal{C}}_k.
\label{eq:stable_canonical_contact}
\end{equation}
The stable canonical contacts are transformed back to each frame within the
segment using the estimated object pose, yielding frame-wise contact targets
$\mathbf{x}^{\mathrm{stab}}_{t,i}$.

\paragraph{Contact-Consistent HOI Optimization.}
We use the stable object-side contacts to refine the human hand trajectory.
Let
$\mathbf{v}^{h}_{t,i}
=
\mathbf{v}^{h}_{i}
(
\boldsymbol{\theta}_t,
\boldsymbol{\tau}_t,
\boldsymbol{\beta}
)$
denote the position of the $i$-th hand vertex under the optimized MANO
articulation and global transformation. The contact-consistency loss is
defined as
\begin{equation}
\mathcal{L}_{\mathrm{contact}}^{h,(t)}
=
\begin{cases}
\displaystyle
\frac{1}{
\left|
\widehat{\mathcal{C}}_{k(t)}
\right|
}
\sum_{i\in\widehat{\mathcal{C}}_{k(t)}}
\left\|
\mathbf{v}^{h}_{t,i}
-
\mathbf{x}^{\mathrm{stab}}_{t,i}
\right\|_2^2,
&
\left|
\widehat{\mathcal{C}}_{k(t)}
\right|>0,
\\[3mm]
0,
&
\text{otherwise}.
\end{cases}
\label{eq:hoi_contact_consistency}
\end{equation}
This loss encourages each stable hand contact vertex to remain close to its
corresponding object-side contact throughout the temporal segment. The total
contact-consistency loss is accumulated over the sequence as
$\mathcal{L}_{\mathrm{contact}}^{h}
=\sum_{t=1}^{T}\mathcal{L}_{\mathrm{contact}}^{h,(t)}$.

To penalize hand--object penetration, we define the signed distance to the
posed object at frame $t$ as
\begin{equation}
d_{o,t}(\mathbf{v})
=
d_o
\left(
\mathbf{R}_{o,t}^{\top}
\left(
\mathbf{v}
-
\mathbf{p}_{o,t}
\right)
\right),
\label{eq:posed_object_sdf}
\end{equation}
where $d_o(\cdot)$ is the signed distance function of the canonical object,
defined to be positive outside and negative inside. The penetration loss over
the complete sequence is
\begin{equation}
\mathcal{L}_{\mathrm{sdf}}^{h}
=
\sum_{t=1}^{T}
\frac{1}{|\mathcal{V}^{h}_t|}
\sum_{\mathbf{v}^{h}_{t,i}\in\mathcal{V}^{h}_t}
\max
\left(
0,
-
d_{o,t}
\left(
\mathbf{v}^{h}_{t,i}
\right)
\right).
\label{eq:hoi_sdf_loss}
\end{equation}
We additionally use a regularization term
$\mathcal{L}_{\mathrm{reg}}^{h}$ that anchors the optimized hand
articulation to its initial estimate and suppresses temporal variation in
both the global hand motion and MANO articulation.

Keeping the estimated object trajectory fixed, we jointly optimize the hand
articulation and global transformations over the complete sequence:
\begin{equation}
\min_{\{\boldsymbol{\theta}_t,\boldsymbol{\tau}_t\}_{t=1}^{T}}
\quad
\mathcal{L}_{\mathrm{HOI}}
=
\lambda_{c}^{h}\mathcal{L}_{\mathrm{contact}}^{h}
+
\lambda_{\mathrm{sdf}}^{h}\mathcal{L}_{\mathrm{sdf}}^{h}
+
\lambda_{\mathrm{reg}}^{h}\mathcal{L}_{\mathrm{reg}}^{h}.
\label{eq:hoi_total_objective}
\end{equation}
The optimized sequence provides a temporally coherent and contact-consistent
human HOI trajectory, together with stable object-side contacts for the
subsequent retargeting module.

\subsection{Contact-Interaction-Preserving Retargeting}
\label{sec:retargeting}

Given the reconstructed human HOI trajectory, we first obtain an initial
dexterous-hand trajectory using keypoint-based retargeting. We then refine this
trajectory through two complementary objectives. Stable contact optimization
explicitly transfers task-relevant object-side contacts, while Laplacian
interaction optimization preserves the local geometric
relationships between the hand and object. The trajectory is finally refined
via residual reinforcement learning in simulation to obtain the executable
result.

\paragraph{Stable Contact Optimization.}
Keypoint-based retargeting reproduces the overall hand configuration but does
not guarantee that task-relevant contacts remain on their demonstrated object
regions. We therefore reuse the stable object-side contacts recovered by the
HOI reconstruction module as explicit retargeting targets.

Let $\mathcal{K}$ denote the set of dexterous-hand fingers, and let $i_f$
denote the human hand vertex associated with finger $f\in\mathcal{K}$. The
active contact fingers at frame $t$ are
\begin{equation}
\mathcal{K}^{c}_t
=
\left\{
f\in\mathcal{K}
\;\middle|\;
i_f\in\widehat{\mathcal{C}}_{k(t)}
\right\},
\label{eq:active_contact_fingers}
\end{equation}
where $\widehat{\mathcal{C}}_{k(t)}$ is the stable contact set obtained from
the preceding module. Let $\mathbf{y}^{d}_{t,f}$ denote the contact position
of dexterous-hand finger $f$. We define the stable contact loss as
\begin{equation}
\mathcal{L}_{\mathrm{contact}}^{\mathrm{ret}}
=
\sum_{t=1}^{T}
\frac{1}{\max(1,|\mathcal{K}^{c}_t|)}
\sum_{f\in\mathcal{K}^{c}_t}
\left\|
\mathbf{y}^{d}_{t,f}
-
\mathbf{x}^{\mathrm{stab}}_{t,i_f}
\right\|_2^2.
\label{eq:stable_contact_retargeting_loss}
\end{equation}
This objective explicitly aligns the dexterous-hand contacts with the stable
object-side contacts while preserving their establishment and release over
time.

\paragraph{Laplacian Interaction Optimization.}
Inspired by the interaction mesh~\cite{ho2010spatial} and its recent
application in OmniRetarget~\cite{yang2025omniretarget}, which preserves
robot--object spatial
relationships in humanoid loco-manipulation, we represent
the demonstrated hand--object interaction as a volumetric graph and preserve
its local structure through Laplacian coordinates.
Let $\mathbf{H}^{h}_t\in\mathbb{R}^{N_h\times3}$ denote the human hand
keypoints selected to correspond to the keypoints of the target dexterous hand.
We sample a fixed set of canonical object points
$\{\bar{\mathbf{o}}_j\}_{j=1}^{N_o}$ and transform them to frame $t$ using
the estimated object pose, yielding
$\mathbf{O}_t
=
[
\mathbf{R}_{o,t}\bar{\mathbf{o}}_j
+
\mathbf{p}_{o,t}
]_{j=1}^{N_o}$.
For the target dexterous hand, let $\mathbf{q}_t$ denote its joint
configuration and $(\mathbf{R}^{d}_t,\mathbf{p}^{d}_t)$ its wrist pose. Its
keypoints $\mathbf{H}^{d}_t$ are obtained through forward kinematics. We
concatenate the human and dexterous-hand keypoints, respectively, with the
same set of object points:
\begin{equation}
\mathbf{V}^{h}_t
=
\begin{bmatrix}
\mathbf{H}^{h}_t\\
\mathbf{O}_t
\end{bmatrix},
\qquad
\mathbf{V}^{d}_t
=
\begin{bmatrix}
\mathbf{H}^{d}_t\\
\mathbf{O}_t
\end{bmatrix}.
\label{eq:interaction_vertex_sets}
\end{equation}
We construct a three-dimensional Delaunay graph over
$\mathbf{V}^{h}_t$ to represent the demonstrated local interaction structure.
Let $\mathcal{N}_t(i)$ denote the neighbors of vertex $i$, and let $w_{t,ij}$
denote the normalized distance-based weight of edge $(i,j)$, satisfying
$\sum_{j\in\mathcal{N}_t(i)}w_{t,ij}=1$. The Laplacian coordinate of vertex
$i$ is defined as
\begin{equation}
\boldsymbol{\delta}_{t,i}(\mathbf{V})
=
\mathbf{v}_{t,i}
-
\sum_{j\in\mathcal{N}_t(i)}
w_{t,ij}\mathbf{v}_{t,j}.
\label{eq:laplacian_coordinate}
\end{equation}
The graph topology and edge weights constructed from the human interaction are
then applied to $\mathbf{V}^{d}_t$. The Laplacian interaction loss is
\begin{equation}
\mathcal{L}_{\mathrm{Lap}}^{\mathrm{ret}}
=
\sum_{t=1}^{T}
\sum_{i=1}^{N_h+N_o}
\left\|
\boldsymbol{\delta}_{t,i}
\left(
\mathbf{V}^{d}_t
\right)
-
\boldsymbol{\delta}_{t,i}
\left(
\mathbf{V}^{h}_t
\right)
\right\|_2^2.
\label{eq:laplacian_interaction_loss}
\end{equation}
By matching local interaction coordinates rather than absolute keypoint
positions, this objective preserves the demonstrated hand--object geometry
while accommodating morphological differences between the human and
dexterous hands.

\paragraph{Trajectory Optimization.}
We jointly optimize the stable contact and Laplacian interaction objectives.
To improve physical feasibility, we additionally use an SDF-based collision
loss $\mathcal{L}_{\mathrm{pene}}^{\mathrm{ret}}$ to penalize hand--object
penetration and self-collision, together with a smoothness loss
$\mathcal{L}_{\mathrm{smooth}}^{\mathrm{ret}}$ to regularize the temporal
variation of the dexterous-hand configuration and wrist pose. Starting from
the keypoint-retargeted trajectory, we optimize
\begin{equation}
\begin{aligned}
\min_{\{
\mathbf{q}_t,
\mathbf{R}^{d}_t,
\mathbf{p}^{d}_t
\}_{t=1}^{T}}
\quad
\mathcal{L}_{\mathrm{ret}}
={}&
\lambda_{\mathrm{Lap}}
\mathcal{L}_{\mathrm{Lap}}^{\mathrm{ret}}
+
\lambda_{\mathrm{contact}}
\mathcal{L}_{\mathrm{contact}}^{\mathrm{ret}}
\\
&+
\lambda_{\mathrm{pene}}
\mathcal{L}_{\mathrm{pene}}^{\mathrm{ret}}
+
\lambda_{\mathrm{smooth}}
\mathcal{L}_{\mathrm{smooth}}^{\mathrm{ret}},
\end{aligned}
\label{eq:retarget_total_loss}
\end{equation}
subject to the joint limits
$\mathbf{q}_{\min}
\leq
\mathbf{q}_t
\leq
\mathbf{q}_{\max}$.

\paragraph{RL-Based Trajectory Refinement.}
To further correct remaining kinematic and dynamics errors, we adopt the
residual policy-learning framework of
ManipTrans~\cite{li2025maniptrans}, which instantiates the residual
learning
paradigm~\cite{silver2018residual,johannink2019residual,alakuijala2021residual,haldar2023teach,ankile2025residualoffpolicyrlfinetuning}
for dexterous manipulation transfer, to refine the optimized trajectory in
physics simulation (Isaac Gym~\cite{makoviychuk2021isaac}).
The trajectory serves as the motion reference for a residual policy trained
with proximal policy optimization
(PPO)~\cite{schulman2017proximalpolicyoptimizationalgorithms}, whose reward
encourages object-motion tracking, dexterous-hand tracking, contact
consistency, and temporal smoothness. Rolling out the learned residual
corrections yields the final executable dexterous-hand trajectory. Note that
the policy is used only to produce this trajectory; no online policy or
visual feedback is required at deployment.

\section{Experiments}

\subsection{Experimental Setup}
\label{sec:experimental_setup}

We evaluate C2Dex from four complementary perspectives: end-to-end
dexterous-hand--object trajectory generation, HOI reconstruction, retargeting
accuracy, and real-world trajectory replay. The first three evaluations are
conducted on benchmark datasets and the last on a real robotic system.

\noindent\textbf{End-to-End Trajectory Generation.}
We compare C2Dex with two video-to-dexterous-manipulation methods,
Do As I Do~\cite{paliwal2026doasido} and
DexImit~\cite{mu2026deximit}, on
DexYCB~\cite{chao2021dexycbbenchmarkcapturinghand} and
TACO~\cite{liu2024taco}.
All methods receive the same monocular human demonstration videos and use
their respective complete inference pipelines to produce
dexterous-hand--object trajectories executed in physics simulation, and the
resulting object trajectory of each method is used for evaluation. We
evaluate all methods on the same set of $45$ manipulation sequences 
sampled from DexYCB and $30$ demonstration sequences  sampled from
TACO. Performance is measured by trajectory success rate based on the
absolute trajectory error (ATE) between the simulated object trajectory and
the corresponding ground-truth trajectory. A trajectory is considered
successful if its ATE-Position is no greater than $0.10$~m and its
ATE-Rotation is no greater than $1.00$~rad under the relaxed criterion or
$0.50$~rad under the strict criterion. Incomplete or invalid trajectories
are counted as failures.

\noindent\textbf{HOI Reconstruction.}
We evaluate HOI reconstruction on DexYCB by comparing C2Dex with
HOLD~\cite{fan2023holdcategoryagnostic3dreconstruction},
DiffHOI~\cite{ye2023diffusionguidedreconstructioneverydayhandobject}, and
BIGS~\cite{on2025bigsbimanualcategoryagnosticinteraction}, where all methods
receive the same input videos and use their respective reconstruction
pipelines. We use simulation
displacement (SD, cm) as the primary measure of
physical interaction
stability, defined as the displacement of the object center of mass over
$1{,}000$ simulation steps. Hand-pose accuracy is evaluated using MPJPE and
P-MPJPE, both reported in mm. Object reconstruction quality is measured using
Chamfer distance (CD, cm$^2$) and F-score at a $5$-mm threshold (F@5, \%).

\noindent\textbf{Retargeting Accuracy.}
We evaluate retargeting accuracy on DexYCB and TACO by comparing C2Dex with
DexPilot~\cite{handa2020dexpilot},
AnyTeleop~\cite{qin2023anyteleop}, and
GeoRT~\cite{yin2025geort}. To isolate retargeting performance from HOI
reconstruction errors, all methods receive the same ground-truth human
hand--object trajectories and retarget the human hand motion to the same
Inspire dexterous-hand embodiment. The ground-truth object trajectory is kept
fixed across all methods. We evaluate the generated dexterous-hand
trajectories using three metrics. Contact precision error
$E_{\mathrm{prec}}$ (mm) measures the distance between the retargeted contacts
and the corresponding ground-truth contact regions on the object surface.
Contact alignment error $E_{\mathrm{align}}$ (deg) measures the angular
discrepancy between the retargeted contact normals and the corresponding
ground-truth contact-surface normals. Maximum penetration depth
$D_{\mathrm{pen}}^{\max}$ (mm) measures the most severe hand--object
interpenetration over the complete trajectory.

\noindent\textbf{Real-World Trajectory Replay.}
We conduct a qualitative real-world feasibility study using $24$ monocular
human demonstration videos covering $8$ contact-rich manipulation tasks.
For each demonstration, the trajectory generated by C2Dex is
directly replayed open-loop on a Unitree G1 humanoid equipped with the
Inspire dexterous hand; the wrist trajectory is executed by the G1 arm
through inverse kinematics. Following the protocol of Do As I
Do~\cite{paliwal2026doasido}, the object is manually placed to match its
initial pose in the demonstration. Consistent
with~\cite{paliwal2026doasido}, we assess physical feasibility
qualitatively, examining whether the replayed trajectories reproduce the
intended manipulation behavior while maintaining stable hand--object
interaction.
The human demonstration videos used in this study were recorded by the
authors themselves; no third-party human subjects were involved, and no
personally identifiable information is contained in the data. This
research is therefore exempt from institutional ethics review.

\noindent\textbf{Implementation Details.}
All reconstruction, trajectory optimization, retargeting, and simulation
experiments are conducted on an Ubuntu server equipped with NVIDIA GeForce
RTX 4090 GPUs.
For trajectory optimization and retargeting, we use learning rates of $0.005$ and $0.02$ for $1{,}000$ and $3{,}000$ iterations, respectively, with loss weights set to $\lambda_{c}^{h}=8.0$, $\lambda_{\mathrm{sdf}}^{h}=0.01$, $\lambda_{\mathrm{reg}}^{h}=10.0$ and $\lambda_{\mathrm{Lap}}=500.0$, $\lambda_{\mathrm{contact}}=2.0\times10^{4}$, $\lambda_{\mathrm{pene}}=1.0\times10^{5}$, $\lambda_{\mathrm{smooth}}=1.0$.

\subsection{End-to-End Trajectory Generation Results}

Table~\ref{tab:end_to_end_trajectory} reports the end-to-end trajectory
success rates on DexYCB and TACO. C2Dex achieves $57.78\%$/$55.56\%$ on
DexYCB and $26.67\%$/$23.33\%$ on TACO under the relaxed and strict
criteria, whereas the strongest baselines reach at most $17.78\%$ and
$10.00\%$, with Do As I Do dropping to $6.67\%$ and $0.00\%$ under the strict
threshold. The consistent gains under both rotation thresholds show that
C2Dex more reliably produces valid trajectories whose executed object motion
agrees with the ground-truth demonstrations across both datasets.

\subsection{HOI Reconstruction Results}
\begin{figure}[t]
    \centering
    \includegraphics[width=\columnwidth]{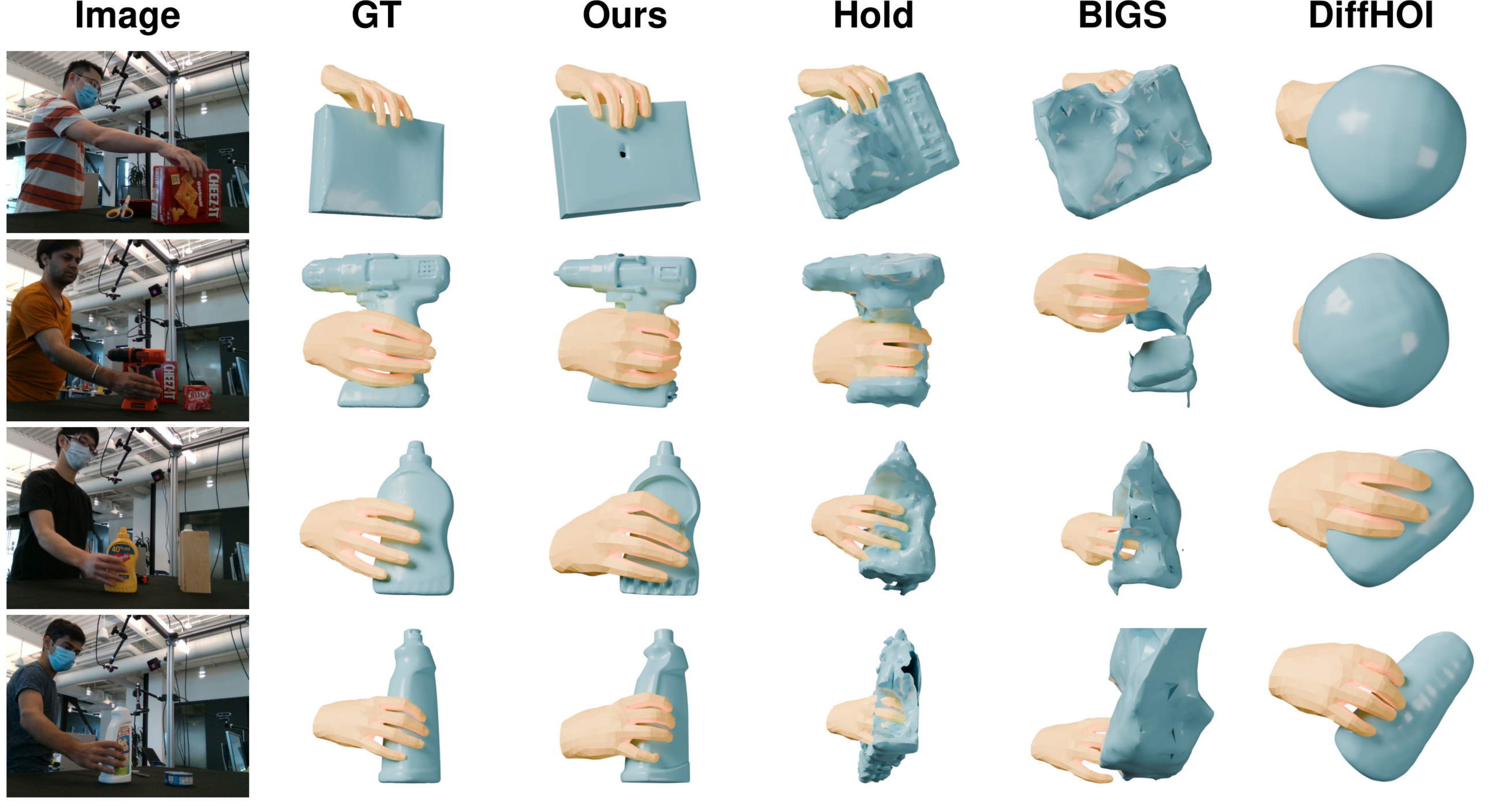}
    \caption{
        Qualitative comparison of HOI reconstruction. C2Dex produces more
        plausible hand--object configurations with more stable contacts and
        less hand--object penetration than the competing methods.
    }
    \label{fig:HOI}
\end{figure}

Table~\ref{tab:quantitative_comparison} summarizes the HOI reconstruction
results on DexYCB. C2Dex achieves the lowest SD of $1.57$~cm, compared with
$4.55$~cm for the best-performing baseline, demonstrating substantially
improved physical stability of the reconstructed hand--object interactions.
C2Dex also obtains the lowest P-MPJPE of $9.27$~mm, indicating more accurate
articulated hand poses after alignment. Although HOLD achieves a lower MPJPE,
this is expected: our contact-consistent optimization prioritizes the
correctness of hand--object contacts over strict fidelity to the initial
per-frame hand pose, and therefore permits deviations in the global hand
placement as long as the demonstrated interaction is preserved. C2Dex further achieves the
best CD and F@5 for object reconstruction. Since the compared methods
employ different object-reconstruction pipelines, these object-side metrics
reflect the quality of the complete reconstruction systems rather than the
isolated effect of contact-consistent refinement.
Fig.~\ref{fig:HOI} shows qualitative comparisons.

\begin{table}[t]
    \centering
    \caption{
        End-to-end dexterous-hand--object trajectory success rates on
        DexYCB and TACO. The translation threshold is $0.10$\,m, with
        rotation thresholds of $1.00$\,rad (relaxed) and $0.50$\,rad
        (strict). Incomplete or invalid trajectories are counted as
        failures.
    }
    \label{tab:end_to_end_trajectory}
    \footnotesize
    \setlength{\tabcolsep}{6.0pt}
    \renewcommand{\arraystretch}{1.15}

    \begin{tabular}{@{}llcc@{}}
        \toprule
        \multirow{2}{*}{Dataset}
        & \multirow{2}{*}{Method}
        & \multicolumn{2}{c}{Trajectory Success Rate (\%) $\uparrow$} \\
        \cmidrule(lr){3-4}
        & & $0.10$\,m / $1.00$\,rad
          & $0.10$\,m / $0.50$\,rad \\
        \midrule

        \multirow{3}{*}{DexYCB}
        & DexImit
        & 11.11 {\scriptsize(5/45)}
        & 11.11 {\scriptsize(5/45)} \\

        & Do As I Do
        & 17.78 {\scriptsize(8/45)}
        & 6.67 {\scriptsize(3/45)} \\

        & \textbf{C2Dex}
        & \textbf{57.78} {\scriptsize(26/45)}
        & \textbf{55.56} {\scriptsize(25/45)} \\

        \midrule

        \multirow{3}{*}{TACO}
        & DexImit
        & 0.00 {\scriptsize(0/30)}
        & 0.00 {\scriptsize(0/30)} \\

        & Do As I Do
        & 10.00 {\scriptsize(3/30)}
        & 0.00 {\scriptsize(0/30)} \\

        & \textbf{C2Dex}
        & \textbf{26.67} {\scriptsize(8/30)}
        & \textbf{23.33} {\scriptsize(7/30)} \\

        \bottomrule
    \end{tabular}
\end{table}

\subsection{Retargeting Results}

\begin{figure}[t]
    \centering
    \includegraphics[width=\columnwidth]{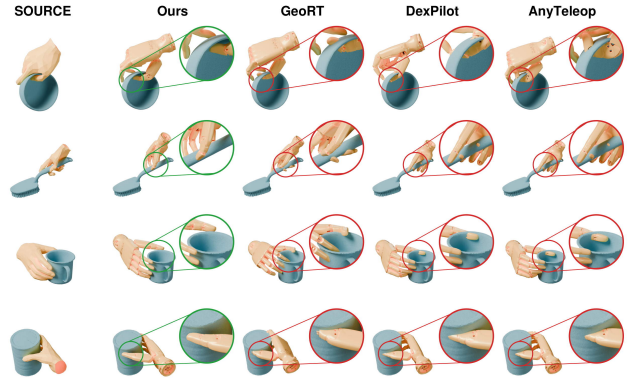}
    \caption{
        Qualitative comparison of retargeting accuracy. C2Dex better retains
        task-relevant contacts while producing substantially less
        hand--object penetration.
    }
    \label{fig:compare_retarget}
\end{figure}

Table~\ref{tab:retarget_gt_all} reports retargeting accuracy using
ground-truth human hand--object trajectories as input. C2Dex consistently
achieves the lowest contact precision error, contact alignment error, and
maximum penetration depth on both datasets. On DexYCB, C2Dex reduces these
errors from the best baseline values of $16.51$~mm, $34.92^\circ$, and
$22.92$~mm to $11.87$~mm, $18.13^\circ$, and $3.99$~mm, respectively, and
similar improvements hold on TACO. The gains in
contact position and normal alignment show that C2Dex more accurately
transfers task-relevant contacts, while the substantial reduction in
penetration demonstrates the benefit of preserving the local
interaction geometry.
Fig.~\ref{fig:compare_retarget} shows qualitative examples.

\subsection{Ablation Studies}
\label{sec:ablation}
We ablate three main components of C2Dex. \emph{w/o CC} replaces cross-frame
contact consistency with frame-wise contact observations. \emph{w/o Lap}
removes the Laplacian interaction loss
$\mathcal{L}_{\mathrm{Lap}}^{\mathrm{ret}}$, and \emph{w/o CPL} removes the
stable contact loss $\mathcal{L}_{\mathrm{contact}}^{\mathrm{ret}}$
(contact-preservation loss).

\begin{table}[t]
    \centering
    \scriptsize
    \setlength{\tabcolsep}{4.0pt}
    \renewcommand{\arraystretch}{1.12}
    \caption{
        HOI reconstruction results on DexYCB. Simulation displacement
        is used as the primary measure of physical interaction quality.
    }
    \label{tab:quantitative_comparison}

    \begin{tabular}{@{}lccccc@{}}
        \toprule
        \multirow{2}{*}{Method}
        & \multicolumn{1}{c}{Physics}
        & \multicolumn{2}{c}{Hand Pose}
        & \multicolumn{2}{c}{Object} \\
        \cmidrule(lr){2-2}
        \cmidrule(lr){3-4}
        \cmidrule(lr){5-6}
        & \shortstack{SD\\(cm) $\downarrow$}
        & \shortstack{P-MPJPE\\(mm) $\downarrow$}
        & \shortstack{MPJPE\\(mm) $\downarrow$}
        & \shortstack{CD\\(cm$^2$) $\downarrow$}
        & \shortstack{F@5\\(\%) $\uparrow$} \\
        \midrule

        HOLD
        & 4.55
        & 13.16
        & \textbf{22.13}
        & 4.64
        & 30.63 \\

        DiffHOI
        & 8.33
        & 16.16
        & 51.11
        & 6.82
        & 33.34 \\

        BIGS
        & 10.60
        & 13.41
        & 23.91
        & 9.09
        & 28.05 \\

        \addlinespace[1pt]

        \textbf{C2Dex}
        & \textbf{1.57}
        & \textbf{9.27}
        & 28.55
        & \textbf{0.33}
        & \textbf{79.65} \\

        \bottomrule
    \end{tabular}
\end{table}

\begin{table}[t]
    \centering
    \footnotesize
    \setlength{\tabcolsep}{6.0pt}
    \renewcommand{\arraystretch}{1.15}
    \caption{
        Retargeting accuracy on DexYCB and TACO using ground-truth human
        hand--object trajectories as input.
    }
    \label{tab:retarget_gt_all}

    \begin{tabular}{@{}llccc@{}}
        \toprule
        Dataset
        & Method
        & \shortstack{$E_{\mathrm{prec}}$\\(mm) $\downarrow$}
        & \shortstack{$E_{\mathrm{align}}$\\(deg) $\downarrow$}
        & \shortstack{$D_{\mathrm{pen}}^{\max}$\\(mm) $\downarrow$} \\
        \midrule

        \multirow{4}{*}{DexYCB}
        & GeoRT
        & 19.47
        & 44.92
        & 28.58 \\

        & DexPilot
        & 16.51
        & 34.92
        & 22.92 \\

        & AnyTeleop
        & 17.49
        & 37.11
        & 23.64 \\

        & \textbf{C2Dex}
        & \textbf{11.87}
        & \textbf{18.13}
        & \textbf{3.99} \\

        \midrule

        \multirow{4}{*}{TACO}
        & GeoRT
        & 13.73
        & 30.19
        & 13.72 \\

        & DexPilot
        & 20.07
        & 39.66
        & 14.58 \\

        & AnyTeleop
        & 21.62
        & 46.67
        & 14.81 \\

        & \textbf{C2Dex}
        & \textbf{11.72}
        & \textbf{18.88}
        & \textbf{5.41} \\

        \bottomrule
    \end{tabular}
\end{table}

Table~\ref{tab:end_to_end_ablation} shows that cross-frame contact consistency
is essential to end-to-end performance: removing it drops the success rate
from $57.78\%$ to $17.78\%$ on DexYCB and from $26.67\%$ to $10.00\%$
(relaxed) on TACO. Removing the Laplacian interaction loss also causes a
large drop on
both datasets, demonstrating the importance of preserving local
hand--object geometry during retargeting. Removing the contact-preservation
loss has a smaller effect under the relaxed criterion, but consistently
degrades performance under the stricter rotation threshold. Together, these
results show that stable contact recovery, local interaction preservation,
and explicit contact transfer play complementary roles in generating robust
dexterous-hand--object trajectories.

\begin{table}[t]
    \centering
    \caption{
        Ablation study on DexYCB and TACO. The evaluation protocol follows
        Table~\ref{tab:end_to_end_trajectory}.
    }
    \label{tab:end_to_end_ablation}
    \footnotesize
    \setlength{\tabcolsep}{5.5pt}
    \renewcommand{\arraystretch}{1.15}

    \begin{tabular}{@{}llcc@{}}
        \toprule
        \multirow{2}{*}{Dataset}
        & \multirow{2}{*}{Variant}
        & \multicolumn{2}{c}{Trajectory Success Rate (\%) $\uparrow$} \\
        \cmidrule(lr){3-4}
        & &
        $0.10$\,m / $1.00$\,rad
        &
        $0.10$\,m / $0.50$\,rad \\
        \midrule

        \multirow{4}{*}{DexYCB}
        & w/o CC
        & 17.78 {\scriptsize(8/45)}
        & 17.78 {\scriptsize(8/45)} \\

        & w/o Lap
        & 26.67 {\scriptsize(12/45)}
        & 20.00 {\scriptsize(9/45)} \\

        & w/o CPL
        & 55.56 {\scriptsize(25/45)}
        & 44.44 {\scriptsize(20/45)} \\

        & \textbf{Full}
        & \textbf{57.78} {\scriptsize(26/45)}
        & \textbf{55.56} {\scriptsize(25/45)} \\

        \midrule

        \multirow{4}{*}{TACO}
        & w/o CC
        & 10.00 {\scriptsize(3/30)}
        & 0.00 {\scriptsize(0/30)} \\

        & w/o Lap
        & 13.33 {\scriptsize(4/30)}
        & 13.33 {\scriptsize(4/30)} \\

        & w/o CPL
        & 23.33 {\scriptsize(7/30)}
        & 13.33 {\scriptsize(4/30)} \\

        & \textbf{Full}
        & \textbf{26.67} {\scriptsize(8/30)}
        & \textbf{23.33} {\scriptsize(7/30)} \\

        \bottomrule
    \end{tabular}
\end{table}

\begin{figure*}[t]
    \centering
    \includegraphics[width=\textwidth]{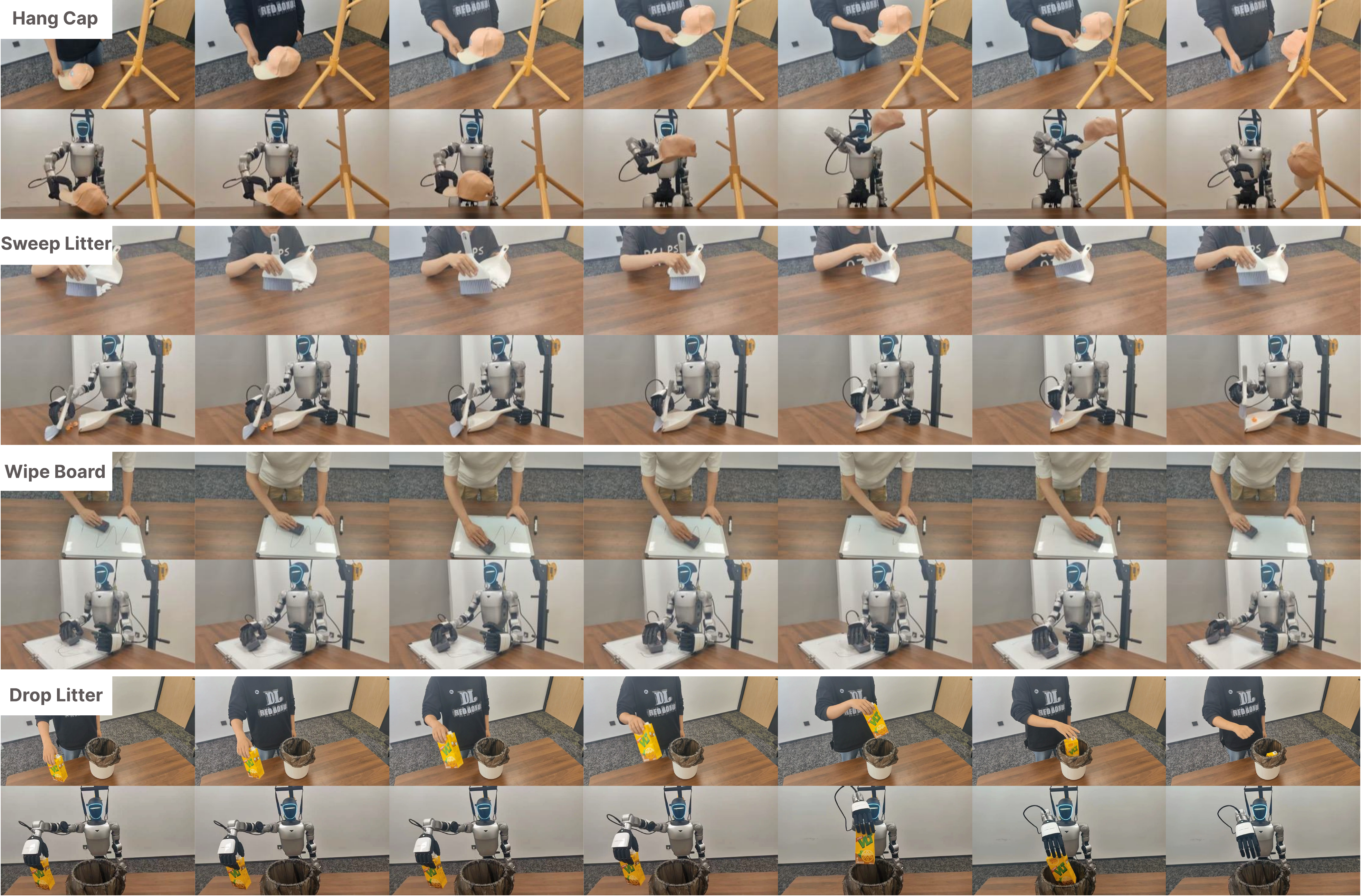}
    \caption{
        Representative real-world trajectory replay results on four
        contact-rich tasks: \emph{Hang Cap}, \emph{Sweep Litter},
        \emph{Wipe Board}, and \emph{Drop Litter}. For each task, the upper
        row shows the human demonstration and the lower row shows the
        corresponding dexterous-hand execution generated by C2Dex.
    }
    \label{fig:task_flow}
\end{figure*}

\subsection{Real-World Trajectory Replay}

Fig.~\ref{fig:task_flow} visualizes the temporal execution of four
representative tasks: \emph{Hang Cap}, \emph{Sweep Litter},
\emph{Wipe Board}, and \emph{Drop Litter}. Across the evaluated tasks, the
replayed trajectories reproduce
the principal manipulation motions and contact transitions observed in the
human demonstrations, demonstrating that the generated trajectories are
physically executable on real hardware under controlled, open-loop replay
conditions. Qualitative results for all eight tasks are available on our
project page.

\section{Conclusion}

We have presented C2Dex, a unified framework that converts monocular human videos into executable dexterous manipulation trajectories via contact-consistent HOI reconstruction and interaction-preserving retargeting. Experiments on DexYCB and TACO demonstrate consistent improvements over state-of-the-art baselines in success rate, reconstruction stability, and retargeting accuracy, and real-robot deployment further validates the physical feasibility of the generated trajectories. Several limitations remain. First, the extraction of hand–object interaction cues is not always reliable under severe occlusion, extreme viewpoints, or objects with intricate geometry. Second, our stable contact representation assumes accurate object pose estimation; pose drift directly corrupts the canonical-space contact aggregation. Third, our method does not yet support complex in-hand manipulation, such as finger gaiting, which involves frequent contact switching beyond our current formulation. In future work, we will improve the robustness of interaction perception under challenging visual conditions and extend our framework toward in-hand manipulation.


\bibliographystyle{IEEEtran}
\bibliography{references}
\end{document}